\documentclass[11pt,a4paper]{article}

\usepackage{iftex}
\ifPDFTeX
  \usepackage[T1]{fontenc}
  \usepackage[utf8]{inputenc}
  \usepackage{lmodern}
\else
  \usepackage{fontspec}
  \defaultfontfeatures{Ligatures=TeX}
\fi

\usepackage{amsmath,amssymb}
\usepackage{microtype}
\usepackage[hyphens]{url}

\usepackage[a4paper,margin=1in]{geometry}

\usepackage{graphicx}
\graphicspath{{figures/}}
\usepackage{booktabs}
\usepackage{array}
\usepackage{tabularx}
\usepackage{longtable}

\usepackage{caption}
\usepackage{titlesec}
\titleformat{\section}{\Large\bfseries}{\thesection}{0.6em}{}
\titleformat{\subsection}{\large\bfseries}{\thesubsection}{0.6em}{}
\titleformat{\subsubsection}{\normalsize\bfseries}{\thesubsubsection}{0.6em}{}
\titlespacing*{\section}{0pt}{1.4ex plus .2ex}{0.7ex plus .2ex}
\titlespacing*{\subsection}{0pt}{1.1ex plus .2ex}{0.5ex plus .2ex}

\usepackage[dvipsnames]{xcolor}
\usepackage[colorlinks=true,
            linkcolor=black,
            citecolor=blue!55!black,
            urlcolor=blue!55!black]{hyperref}
\hypersetup{
  pdftitle={KAYRA: A Microservice Architecture for AI-Assisted Karyotyping with Cloud and On-Premise Deployment},
  pdfauthor={Attila Pintér; Javier Rico; Attila Répai; Jalal Al-Afandi; Adrienn Éva Borsy; András Kozma; Hajnalka Andrikovics; György Cserey}
}

\renewcommand{\arraystretch}{1.15}

\renewenvironment{abstract}
  {\begin{center}\bfseries Abstract\end{center}\vspace{-0.6em}\begin{quote}\small}
  {\end{quote}\vspace{0.2em}}

\title{KAYRA: A Microservice Architecture for AI-Assisted Karyotyping\\
with Cloud and On-Premise Deployment}

\author{Attila Pintér\(^{1,2}\), Javier Rico\(^{1}\), Attila Répai\(^{1}\), Jalal Al-Afandi\(^{1}\),\\
Adrienn Éva Borsy\(^{3}\), András Kozma\(^{3}\), Hajnalka Andrikovics\(^{3}\), György Cserey\(^{2}\)}

\date{}

\begin{document}
\maketitle

\begin{abstract}
We present KAYRA, an end-to-end karyotyping system that operates inside the operational constraints of a clinical cytogenetic laboratory. KAYRA is architected as a containerized microservice pipeline whose ML stack combines an EfficientNet-B5 + U-Net semantic segmenter, a Mask R-CNN (ResNet-50 + FPN) instance detector, and a ResNet-18 classifier, orchestrated through a cascaded ROI-narrowing strategy that focuses each downstream model on the chromosome-bearing region. The same container images are deployed both as a cloud service and as an on-premise installation, supporting clinical environments where patient-data egress is not permitted as well as those where it is.

A pilot clinical evaluation against two commercial reference karyotyping systems on 459 chromosomes from 10 metaphase spreads shows segmentation accuracy of 98.91~\% (vs.~78.21~\% / 40.52~\%), classification accuracy of 89.1~\% (vs.~86.9~\% / 54.5~\%), and rotation accuracy of 89.76~\% (vs.~94.55~\% / 78.43~\%). KAYRA improves over the older density-thresholding reference on all three axes (\(p < 0.0001\) for segmentation and classification by Fisher's exact test on chromosome-level counts), and on segmentation also against the modern AI-supported reference (\(p < 0.0001\)); on classification the difference vs.~the modern AI reference is not statistically significant at the present test-set size (\(p = 0.34\)). The system reaches TRL~6 maturity and integrates the human-in-the-loop expert-review workflow that diagnostic cytogenetic practice requires.

The thesis of this paper is that a multi-model cytogenetic AI service can be packaged as a microservice architecture supporting flexible deployment --- cloud-hosted or on-premise --- while delivering strong empirical performance on a pilot clinical evaluation.

\smallskip
\noindent\emph{Keywords:} karyotyping, instance segmentation, microservice architecture, on-premise deployment, clinical AI, Mask R-CNN, U-Net.
\end{abstract}

\noindent
\(^{1}\) Jedlik Innovation Ltd., Budapest, Hungary \quad
\(^{2}\) Pázmány Péter Catholic University, Faculty of Information Technology and Bionics, Budapest, Hungary \quad
\(^{3}\) Central Hospital of Southern Pest --- National Institute of Hematology and Infectious Diseases, Laboratory of Molecular Genetics, Budapest, Hungary

\noindent
Correspondence: \texttt{pinter.attila@itk.ppke.hu}, \texttt{cserey.gyorgy@itk.ppke.hu}

\bigskip\hrule\bigskip

\section{1. Introduction}

Karyotyping --- the visual identification, ordering, and pairing of human chromosomes from G-banded metaphase microscope images --- is a foundational clinical-diagnostic procedure. It underpins prenatal screening (trisomy 13/18/21, sex-chromosome aneuploidies), oncohematology (the Philadelphia translocation \texttt{t(9;22)(q34;q11)} in chronic myeloid leukaemia), and hereditary-disease workups. In current practice the procedure is performed manually, on average requires 30--90~minutes per case, and depends on a small population of highly trained cytogeneticists; trainee onboarding spans many months. It is the rate-limiting step in many clinical cytogenetic laboratories.

Recent deep-learning work has shown that segmentation and classification accuracy on metaphase images can match or exceed manual results. The 2023 AutoKary2022 benchmark [You et al.~2023] released a 612-image, \textasciitilde27\,000-instance dataset and established Mask R-CNN-style pipelines as a strong baseline; the 2024 KaryoXpert framework [Xia et al.~2024] reaches 94.09~\% accuracy on 6\,592 clinical samples without per-pixel mask labels and is deployed clinically; DaCSeg [Lin et al.~2023] addresses overlapping-instance failure modes via a divide-and-conquer wrapper. The published literature is rich on architectures and benchmarks.

What we found less treated is the system-engineering problem: how does one package a multi-model karyotyping pipeline --- typically a sequence of detection, segmentation-refinement, duplicate-resolution, and classification networks --- into a deployable clinical service that runs reliably within the operational and infrastructural constraints of a cytogenetic laboratory? In our experience working with partner laboratories, an on-premise install option is often a precondition for evaluation due to patient-data residency requirements, so dual cloud/on-premise deployability becomes a binding design constraint rather than a downstream packaging concern. This paper presents KAYRA, a deployed karyotyping system that meets this dual-deployment constraint and shows strong empirical performance on a pilot clinical evaluation. Our contributions are:

\begin{enumerate}
  \item A microservice architecture (Section 3) decomposing the karyotyping computation into single-responsibility containerized services --- a pipeline orchestrator, four model services (semantic segmentation, instance detection, duplicate-resolution, classification + rotation), and a backend that handles ingest, asynchronous task queueing, interactive editing, and karyogram synthesis. The same container images are used for cloud and on-premise deployment.
  \item A cascaded ROI-narrowing pipeline (Section 4) that uses a fast Otsu + connected-components stage to crop the metaphase region at original resolution, then a U-Net stage to refine the chromosome-bearing area at a stable downstream-friendly resolution, and only then runs Mask R-CNN on the narrowed region. The cascade lets each downstream model see only the region it can process well, in line with divide-and-conquer principles motivated by recent overlap-segmentation work [Lin et al.~2023].
  \item A clinical validation (Section 5) on 459 chromosomes from 10 metaphase spreads against two commercial reference systems, reporting segmentation, classification, and chromosome-rotation accuracy.
\end{enumerate}

\section{2. Related Work}

We position KAYRA against three threads of recent work.

\emph{2.1 End-to-end chromosome segmentation and classification.} The AutoKary2022 dataset [You et al.~2023] provides 612 G-banded metaphase images with dense polygon annotations across 50 patients and benchmarks Mask R-CNN-style pipelines as a strong baseline. KaryoXpert [Xia et al.~2024] combines morphological algorithms with deep models, achieving 94.09~\% accuracy on 6\,592 clinical samples without per-pixel mask labels, and is deployed clinically; it is the closest deployed analogue to KAYRA in the published literature, but its emphasis is label-efficiency rather than service architecture. DaCSeg [Lin et al.~2023] explicitly addresses overlap with a divide-and-conquer wrapper, observing that classical Mask R-CNN-based methods cannot handle chromosome overlap without help.

\emph{2.2 Backbones and architectures.} Our deployed pipeline composes three foundational architectures: Mask R-CNN [He et al.~2017] (with the Faster R-CNN [Ren et al.~2015] region-proposal backbone) for instance segmentation; U-Net [Ronneberger et al.~2015] for the semantic-segmentation refinement stage; and a ResNet-18 [He et al.~2016] for chromosome classification. We use these particular models because they are well supported by mature deployment tooling; transformer-based alternatives such as Mask2Former [Cheng et al.~2022] (with Swin Transformer backbones [Liu et al.~2021]) are an active research direction (Section 6) but not the deployed configuration.

\emph{2.3 Cascade design and multi-model pipelines.} Cascade R-CNN [Cai \& Vasconcelos 2018] popularises the cascade-of-detectors principle that motivates our cascaded-ROI strategy (Section 4), although our cascade is heterogeneous --- Otsu pre-crop, U-Net semantic refiner, Mask R-CNN instance segmenter, classifier --- rather than a homogeneous IoU-threshold cascade. What we found less developed in the published literature is the system-engineering treatment of multi-model cytogenetic AI: how to compose these backbones into a deployable clinical service whose installation profile matches what laboratories actually have available. That is the gap KAYRA targets.

\section{3. System Overview}

KAYRA decomposes the karyotyping computation into a small number of containerized microservices that communicate over HTTP/REST. The decomposition follows the single-responsibility principle: a CPU-side pipeline orchestrator that coordinates the per-image flow; four model services (semantic segmentation, instance detection, duplicate-/overlap-resolution, classification + rotation); a backend that handles user-facing concerns (multi-tenant identity, image ingest with format normalization, an asynchronous task queue, the interactive editing API, and final karyogram synthesis with ISCN-suggestion generation); and a web frontend that supports interactive expert review. Each service is independently deployable, scalable, and updatable, and the orchestrator implements degraded operation: if an upstream service is unreachable, it skips that stage and proceeds with the partial result rather than failing the whole pipeline. This is a deliberate clinical-AI design choice --- partial output that the cytogeneticist can correct is more useful than no output at all.

The same container images run in two deployment configurations. A cloud configuration places each model service on a managed Kubernetes platform with separate GPUs and standard managed-infrastructure components (autoscaling, ingress, secret management, identity). An on-premise configuration runs the same images under Docker Compose with all GPU services pinned to a shared device. The choice between the two is operational, not architectural: cloud installations gain managed-infrastructure benefits, while on-premise installations gain data-residency compliance and local clinical-IT integration. In both configurations, multi-tenancy at the data layer separates institutional data, and an audit trail records every interactive correction the cytogeneticist applies.

\begin{figure}[t]
  \centering
  \begin{minipage}[b]{0.46\linewidth}
    \centering
    \includegraphics[width=\linewidth]{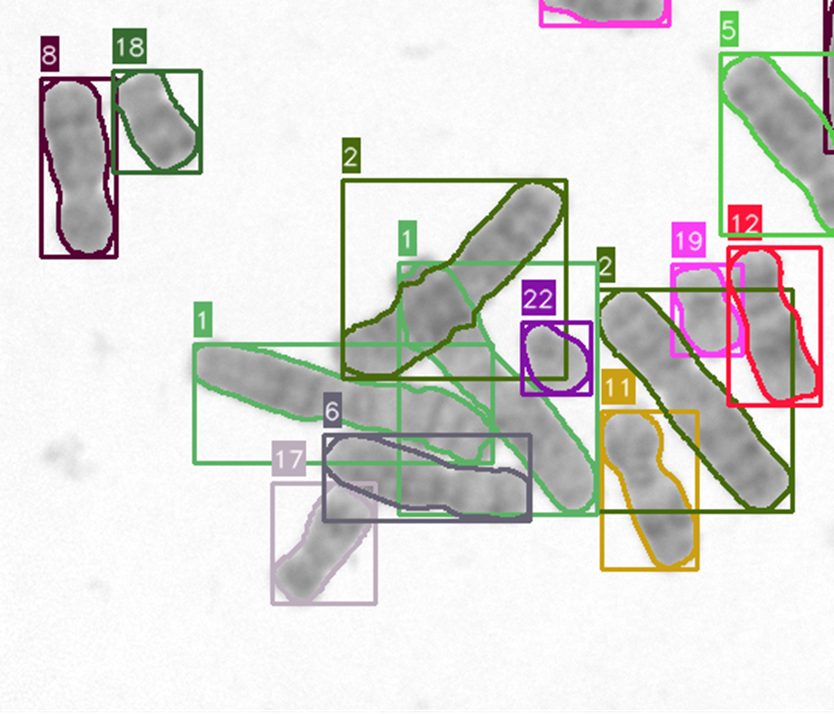}\\[2pt]
    {\small (a) Mask R-CNN bounding-box detections}
  \end{minipage}\hfill
  \begin{minipage}[b]{0.46\linewidth}
    \centering
    \includegraphics[width=\linewidth]{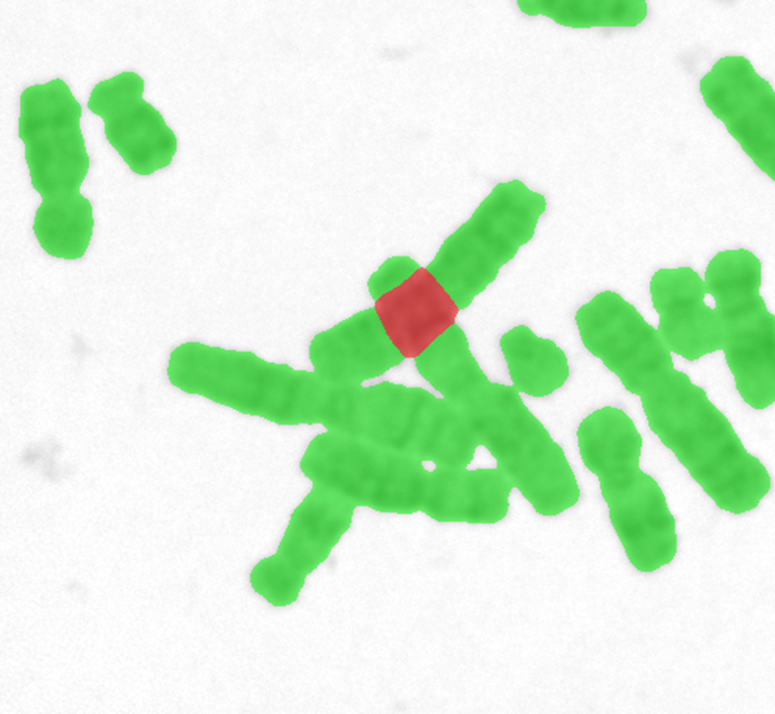}\\[2pt]
    {\small (b) Per-instance segmentation masks}
  \end{minipage}
  \caption{Cooperation between instance detection and segmentation on a cluster of crossing chromosomes. Panel~(a) shows Mask R-CNN's per-chromosome bounding-box proposals --- including overlapping, ambiguous boxes around the crossing region. Panel~(b) shows the corresponding per-instance segmentation masks: the U-Net-refined ROI feeds Mask R-CNN's mask head, which separates the crossing chromosomes (highlighted contour) from their neighbours. This division of labour between detection (proposing where) and segmentation (deciding which pixels belong to which instance) is the case that the cascaded ROI design of Section 4 is built to handle.}
  \label{fig:det-seg-cooperation}
\end{figure}

The web frontend implements the human-in-the-loop expert-review workflow: cytogeneticists can visualise the automatic segmentation and classification, then delete, merge, split, redraw, reclassify, rotate, or flip individual chromosomes before signing off the final karyogram. Every correction is audited and versioned, so the history of an image is fully reconstructable. Figure~\ref{fig:det-seg-cooperation} illustrates how the instance detector and the segmentation stage cooperate on a cluster of crossing chromosomes --- the case that motivates the cascaded ROI design of Section 4. A black-box REST integration-test suite covering the upload~$\to$ segmentation~$\to$ editing~$\to$ karyogram path gates every release with statistical quality checks --- IoU comparison against ground-truth annotations and per-class precision/recall on a held-out validation corpus.

\section{4. ML Pipeline and Cascaded ROI}

\subsection{4.1 Eight-stage flow}

For each metaphase image the pipeline orchestrator runs the following sequence:

\begin{enumerate}
  \item \emph{Otsu-prefilter crop.} Downscale the original (typically 1830$\times$1830) to a 256-pixel max-dimension thumbnail; binarise with Otsu's method [Otsu 1979]; run connected-components labelling, filtering by minimum object area; union the surviving bounding boxes; rescale the union back to original resolution and apply a small margin. Result: \texttt{crop1}, the original-resolution chromosome region.
  \item \emph{Downscale + edge-pad to 992$\times$992.} \texttt{crop1} is downscaled with \texttt{min\_dim=512} and \texttt{max\_dim=992} constraints, then padded with edge-replication to a fixed 992$\times$992 input.
  \item \emph{Semantic segmentation.} An EfficientNet-B5 + U-Net network produces a three-class mask (\texttt{background}, \texttt{chromosome}, \texttt{overlap}).
  \item \emph{Mask-bbox crop.} The non-zero region of the semantic mask in \texttt{crop1} coordinates is bounded with a small margin and converted into \texttt{crop2}, a tighter ROI at original resolution.
  \item \emph{Instance segmentation.} Mask R-CNN (ResNet-50 + FPN) is run on \texttt{crop2} at two angles (0\textdegree{} and 45\textdegree{}) to handle near-rotation invariance, producing per-chromosome bounding boxes and pixel-precise masks; results are merged.
  \item \emph{Duplicate / overlap resolution.} A dedicated stage merges close-by detections under tunable thresholds and refines the final per-chromosome mask using the upscaled semantic mask as a sanity check.
  \item \emph{Classification + rotation.} Each detection is cropped with its mask and passed to a ResNet-18 classifier outputting a class probability vector and a sin/cos rotation-angle estimate. Patches landing in the \texttt{Unknown} class go through a second classifier pass with stronger augmentation.
  \item \emph{Coordinate back-transformation and karyogram assembly.} Annotations are mapped from \texttt{crop2}~$\to$ \texttt{crop1}~$\to$ original-image coordinates and emitted as JSON; the backend composes the ISCN-grouped karyogram image (groups \texttt{1--3, 4--5, 6--12, 13--15, 16--18, 19--22, X, Y, Unknown}).
\end{enumerate}

\subsection{4.2 Why the cascade matters}

The cascade has three engineering payoffs.

First, the Mask R-CNN model --- the largest in the pipeline --- sees a substantially smaller and tighter input than it would on the full image: empirically \texttt{crop2} is \textasciitilde 1349$\times$1510 vs.~1830$\times$1830, a \textasciitilde 34~\% pixel reduction concentrated on the chromosome-bearing region. The downstream RPN proposal load and ROI-head computation reduce proportionally, and the model is presented with a denser, less-background-dominated crop on which it generalises better.

Second, the semantic-segmentation stage operates on a fixed-size input of 992$\times$992 pixels regardless of the original metaphase resolution. Stable input shape is a long-standing engineering benefit for deployed inference systems --- it stabilises memory profiles, simplifies batching, and lets framework-level caches reuse decisions across requests instead of re-deciding per call.

Third, the cascaded approach mirrors the divide-and-conquer principle motivated by DaCSeg [Lin et al.~2023]: each model in the chain sees only the region it can process well. Mask R-CNN's known weakness on heavily overlapping instances is partially mitigated by feeding it U-Net-refined ROIs, while U-Net's globally-averaged receptive field benefits from the Otsu pre-cropping that removes irrelevant background.

\subsection{4.3 Edge-replication padding}

The 992$\times$992 padding around \texttt{crop1} uses edge-replication, not constant (white-background) padding. The original U-Net paper [Ronneberger et al.~2015] uses mirror-padding in its overlap-tile strategy specifically to avoid border discontinuities; constant padding introduces sharp gradient features at the image boundary that the convolutional receptive field interprets as artefacts. Alguacil et al.~[Alguacil et al.~2021] empirically confirm that constant zero padding is the worst-performing choice among standard alternatives for FCN networks. In the chromosome-imaging setting, peripheral chromosomes are clinically meaningful --- losing one to a border artefact loses information that cannot be recovered downstream --- so we use the cheapest faithful-statistics alternative, edge replication.

\subsection{4.4 Training data}

KAYRA's models were trained on a combined dataset of:

\begin{itemize}
  \item \textasciitilde 24\,000 expert-annotated karyograms from the Cytolab partner laboratory (PHA-stimulated peripheral blood);
  \item 297 PHA-stimulated peripheral blood metaphase triplets (input / karyogram / segmentation) and 430 spontaneous bone-marrow triplets jointly contributed by DPC OHII and PPCU-ITK;
  \item 160 expert-annotated structurally-abnormal karyograms (translocations, deletions, inversions, duplications) for targeted rare-class training;
  \item 60 Philadelphia-positive triplets (\texttt{t(9;22)}) for specific-aberration validation;
  \item additional acquisitions from two commercial cytogenetic platforms (\textasciitilde 145 cases) for cross-vendor robustness.
\end{itemize}

Clinically-sourced files follow a uniform \texttt{\{patient\_id\}\_\allowbreak\{year\}\_\allowbreak\{image\_no\}\_\allowbreak\{cultivation\}\_\allowbreak\{type\}.tif} naming convention that enables automated train/validation/test partitioning by patient identifier without data leakage. Public datasets (BioImLab, CloudDataLab, Coriell) supplement the in-house corpus.

\section{5. Evaluation}

\subsection{5.1 Clinical accuracy}

We evaluated the segmentation, classification, and rotation accuracy of KAYRA against two commercial reference systems on a fixed test set of 10 metaphase spreads containing 459 chromosomes drawn from the DPC OHII routine workflow:

\begin{itemize}
  \item 7 normal karyograms (6 $\times$ 46,XX, 1 $\times$ 46,XY)
  \item 1 $\times$ 45,XX,$-$8
  \item 1 $\times$ 45,XX,$-$10
  \item 1 $\times$ 47,XX,+21
\end{itemize}

Five spreads originated from PHA-stimulated peripheral blood (the technically easier preparation) and five from spontaneous bone-marrow culture (the harder preparation). The 10 evaluation spreads are held out from KAYRA's training and validation sets; train/val/test partitioning is performed at the patient-identifier level to prevent data leakage. Reference systems are referred to anonymously: Reference~1 is a long-established density-thresholding system widely used in routine diagnostics; Reference~2 is a commercial AI-supported system. Ground-truth was determined by senior cytogeneticists blinded to which system produced each output.

The 459 chromosomes are clustered within 10 metaphase spreads (\textasciitilde 46 chromosomes per spread); chromosome-level Fisher's exact tests are reported below for comparability with prior work but should be interpreted with the caveat that observations within a spread are not independent. Per-spread aggregate metrics agree directionally with the chromosome-level results; a properly cluster-aware analysis is part of the planned multi-centre validation (see Section 6).

\paragraph{Segmentation.} KAYRA correctly segments 454~/~459 chromosomes (98.91~\%), with 5 chromosomes (1.09~\%) merged with another touching object and 0 chromosomes missed. Reference~2 reaches 78.21~\% correct, and Reference~1 only 40.52~\%; the difference is statistically significant by Fisher's exact test on chromosome-level counts (\(p < 0.0001\)).

\begin{table}[h]
  \centering
  \small
  \begin{tabular}{lrrr}
    \toprule
    System & Correct & Merged with other object & Missed \\
    \midrule
    KAYRA       & 454 (98.91~\%) & 5 (1.09~\%)   & 0 (0.00~\%)  \\
    Reference 2 & 359 (78.21~\%) & 56 (12.20~\%) & 44 (9.59~\%) \\
    Reference 1 & 186 (40.52~\%) & 273 (59.48~\%) & 0 (0.00~\%) \\
    \bottomrule
  \end{tabular}
  \caption{Segmentation accuracy on 459 chromosomes from 10 metaphase spreads. Fisher's exact test, KAYRA vs.~either reference, \(p < 0.0001\).}
  \label{tab:segmentation}
\end{table}

\paragraph{Vertical orientation.} KAYRA correctly orients 412~/~459 chromosomes (89.76~\%); Reference~2 reaches 94.55~\% and Reference~1 78.43~\%. KAYRA is significantly better than Reference~1 (\(p < 0.0001\)) but slightly behind Reference~2 --- an honest area for further work, likely addressable by enriching the rotation-augmentation distribution during training.

\paragraph{Classification.} KAYRA correctly classifies 409~/~459 chromosomes (89.1~\%); Reference~2 reaches 86.9~\% and Reference~1 54.5~\%. The chromosome-group breakdown shows the largest improvements over the references on the medium-sized B and C groups; the smaller D, E, F, and G groups are nearly comparable across the three systems.

\begin{table}[h]
  \centering
  \small
  \begin{tabular}{lrr}
    \toprule
    System & Correct classification & Incorrect classification \\
    \midrule
    KAYRA       & 409 (89.1~\%) & 50 (10.9~\%)  \\
    Reference 2 & 399 (86.9~\%) & 60 (13.1~\%)  \\
    Reference 1 & 250 (54.5~\%) & 209 (45.5~\%) \\
    \bottomrule
  \end{tabular}
  \caption{Classification accuracy on 459 chromosomes. KAYRA vs.~Reference~1: \(p < 0.0001\). KAYRA vs.~Reference~2: \(p = 0.34\) (not statistically significant on this test-set size, motivating a planned multi-centre validation).}
  \label{tab:classification}
\end{table}

\begin{figure}[t]
  \centering
  \includegraphics[width=0.92\linewidth]{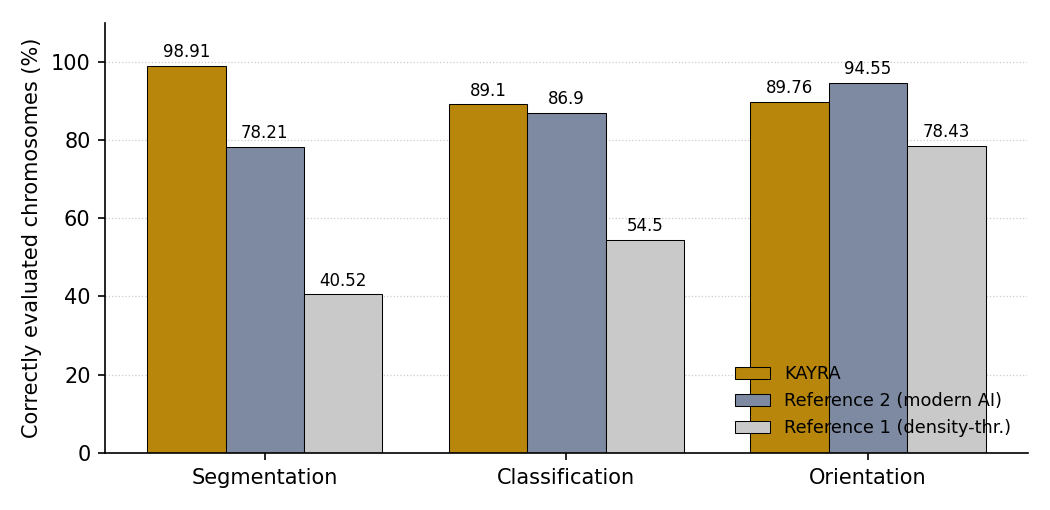}
  \caption{Aggregate accuracy on 459 chromosomes from 10 metaphase spreads, for KAYRA versus the two commercial reference systems. KAYRA improves over the older density-thresholding reference on all three axes (\(p < 0.0001\) for segmentation and classification, by Fisher's exact test on chromosome-level counts) and over the modern AI-supported reference on segmentation (\(p < 0.0001\)); the classification gap to the modern reference is not significant on this test-set size (\(p = 0.34\), see Section 5.1), and the orientation axis remains the one area where KAYRA lags the modern reference.}
  \label{fig:results-bars}
\end{figure}

Figure~\ref{fig:results-bars} summarises the three accuracy axes for the three systems side by side, and Table~\ref{tab:per-class} breaks the classification axis down to per-chromosome-class counts against Reference~1 (the per-class breakdown for Reference~2 was not collected during the present pilot).

\begin{table}[t]
  \centering
  \small
  \caption{Per-class classification on 459 chromosomes from 10 metaphase spreads: KAYRA versus Reference~1 (the older density-thresholding system). For each chromosome class we report the count (and percentage) of correctly classified instances and the chromosome-level Fisher's exact test p-value of KAYRA against Reference~1. Rows with \(p < 0.05\) are bolded. The Y column has only a single instance in this test set and is reported for completeness; no inferential test is meaningful at this sample size.}
  \label{tab:per-class}
  \renewcommand{\arraystretch}{1.05}
  \begin{tabular}{c c r r c}
    \toprule
    Group & Class & KAYRA correct & Reference 1 correct & p-value \\
    \midrule
    A & 1  & 19 / 20 (95.0\%)  & 15 / 20 (75.0\%)  & 0.1818 \\
    A & 2  & 19 / 20 (95.0\%)  & 16 / 20 (80.0\%)  & 0.3416 \\
    A & 3  & 19 / 20 (95.0\%)  & 14 / 20 (70.0\%)  & 0.0915 \\
    B & 4  & 19 / 20 (95.0\%)  & 15 / 20 (75.0\%)  & 0.1818 \\
    \textbf{B} & \textbf{5}  & \textbf{20 / 20 (100\%)} & \textbf{8 / 20 (40.0\%)}  & \textbf{0.0001} \\
    \textbf{C} & \textbf{6}  & \textbf{20 / 20 (100\%)} & \textbf{11 / 20 (55.0\%)} & \textbf{0.0012} \\
    \textbf{C} & \textbf{7}  & \textbf{19 / 20 (95.0\%)} & \textbf{11 / 20 (55.0\%)} & \textbf{0.0084} \\
    \textbf{C} & \textbf{8}  & \textbf{19 / 19 (100\%)}  & \textbf{10 / 19 (52.6\%)} & \textbf{0.0011} \\
    \textbf{C} & \textbf{9}  & \textbf{19 / 20 (95.0\%)} & \textbf{4 / 20 (20.0\%)}  & \textbf{<0.0001} \\
    \textbf{C} & \textbf{10} & \textbf{16 / 19 (84.2\%)} & \textbf{6 / 19 (31.6\%)}  & \textbf{0.0025} \\
    \textbf{C} & \textbf{11} & \textbf{20 / 20 (100\%)} & \textbf{13 / 20 (65.0\%)} & \textbf{0.0083} \\
    \textbf{C} & \textbf{12} & \textbf{19 / 20 (95.0\%)} & \textbf{9 / 20 (45.0\%)}  & \textbf{0.0012} \\
    \textbf{D} & \textbf{13} & \textbf{17 / 20 (85.0\%)} & \textbf{10 / 20 (50.0\%)} & \textbf{0.0407} \\
    \textbf{D} & \textbf{14} & \textbf{20 / 20 (100\%)} & \textbf{7 / 20 (35.0\%)}  & \textbf{<0.0001} \\
    \textbf{D} & \textbf{15} & \textbf{17 / 20 (85.0\%)} & \textbf{9 / 20 (45.0\%)}  & \textbf{0.0187} \\
    \textbf{E} & \textbf{16} & \textbf{17 / 20 (85.0\%)} & \textbf{10 / 20 (50.0\%)} & \textbf{0.0407} \\
    E & 17 & 19 / 20 (95.0\%) & 14 / 20 (70.0\%) & 0.0915 \\
    E & 18 & 17 / 20 (85.0\%) & 12 / 20 (60.0\%) & 0.1552 \\
    F & 19 & 15 / 20 (75.0\%) & 9 / 20 (45.0\%)  & 0.1053 \\
    \textbf{F} & \textbf{20} & \textbf{18 / 20 (90.0\%)} & \textbf{11 / 20 (55.0\%)} & \textbf{0.0310} \\
    G & 21 & 12 / 21 (57.1\%) & 18 / 21 (85.7\%) & 0.0855 \\
    G & 22 & 10 / 20 (50.0\%) & 12 / 20 (60.0\%) & 0.7512 \\
    \textbf{C} & \textbf{X}  & \textbf{19 / 19 (100\%)}  & \textbf{6 / 19 (31.6\%)}  & \textbf{<0.0001} \\
    G & Y  & 0 / 1 (0\%)      & 0 / 1 (0\%)      & --- \\
    \bottomrule
  \end{tabular}
\end{table}

The per-class breakdown supports the aggregate result and adds two qualitative observations. (i) KAYRA's strongest gains over Reference~1 are concentrated on the medium-sized B and C groups (chromosomes 5--12) and on the X chromosome --- exactly the classes where Reference~1's density-thresholding heuristic struggles with overlapping or similarly-sized chromosomes; the chromosome-level Fisher's-exact test on these classes gives \(p < 0.01\) throughout. (ii) The smallest classes (G group: 21, 22) are the only ones where KAYRA does not consistently beat Reference~1; class 22 in particular is a known weakness for current segmentation backbones because the chromosomes are short, lightly banded, and easily confused with nuclear debris. Class 21 in this test set is enriched by the 47,XX,+21 case, which gives Reference~1's heuristic a within-spread template advantage on this specific morphology.

\paragraph{Cultivation type robustness.} A separate analysis on the same 459-chromosome corpus, partitioned by cultivation type, shows KAYRA correctly identifying chromosomes in 93.5~\% of bone-marrow samples versus 86.5~\% of PHA samples. Counter-intuitively, KAYRA performs better on the technically harder bone-marrow preparation, because the training corpus is itself dominated by bone-marrow samples --- a distribution-shift observation that informs our active-learning roadmap.

\subsection{5.2 Integration tests and reliability}

A dedicated black-box REST integration-test suite exercises the full multi-tenant authorisation model and the upload~$\to$ editing~$\to$ karyogram path, including statistical quality gates: Intersection-over-Union mask comparison against ground-truth annotations and scikit-learn classification reports on a 60-image batch. The automated suite's per-class precision/recall is consistent (within 1 percentage point on 22 of 24 chromosome classes) with the manual evaluation of Section 5.1. The Y chromosome is the lowest-recall class, attributable to its under-representation in the training corpus.

\section{6. Limitations}

The following limitations should be noted.

\paragraph{(i) Single-laboratory validation.} The 459-chromosome accuracy benchmark is drawn from a single laboratory. A multi-centre prospective validation with several partner laboratories is in preparation and will supplement the present results.

\paragraph{(ii) Rotation accuracy gap.} KAYRA's rotation accuracy lags Reference~2 by \textasciitilde 5 percentage points (89.76~\% vs.~94.55~\%). We expect this gap to close with rotation-augmentation enrichment during training; the underlying classifier already produces a sin/cos rotation estimate, the issue is the distribution of rotations seen at training time.

\paragraph{(iii) Y-chromosome under-representation.} The Y chromosome is the lowest-recall class in our automated quality reports, attributable to its under-representation in the training corpus. Targeted rare-class augmentation, including synthetic samples generated by the project's separate generative system, is part of the active-learning roadmap.

\paragraph{(iv) Regulatory scope.} The system is currently approved for research and pre-clinical use; CE-marking and FDA SaMD pathway preparation has begun but is not within the scope of the current TRL-6 deployment.

\paragraph{(v) Backbone-agnostic architecture.} The microservice contracts are deliberately model-agnostic: an affected model service can be replaced behind its HTTP contract --- including with transformer-based successors such as Mask2Former [Cheng et al.~2022] --- without disturbing the rest of the pipeline. Such migrations are part of our future-work roadmap.

\section{7. Conclusion}

KAYRA is an end-to-end karyotyping system that takes a metaphase microscope image as input and produces an ISCN-compliant karyogram and ISCN-suggestion code, packaged as containerized microservices that run as both a cloud service and an on-premise institutional deployment. Our principal contributions are: (i) a microservice decomposition with single-responsibility model services and a degradable orchestrator; (ii) a cascaded ROI-narrowing pipeline that lets each downstream model see only the region it can process well; and (iii) a clinical validation against two commercial reference systems on 459 chromosomes from 10 metaphase spreads, showing 98.91~\% segmentation, 89.1~\% classification, and 89.76~\% rotation accuracy.

The thesis of the paper is that a multi-model cytogenetic AI service can be packaged as a microservice architecture supporting flexible deployment --- cloud-hosted or on-premise --- while delivering strong empirical performance on a pilot clinical evaluation and integrating cleanly into the human-in-the-loop expert workflow that diagnostic cytogenetics demands. The architectural choices documented here are backbone-agnostic, and we expect them to remain useful as the upstream model components evolve.

\section*{Acknowledgements}

This work was supported by the Hungarian National Research, Development and Innovation Office (NKFIH) under project number 2021-1.1.4-GYORSÍTÓSÁV-2022-00054, executed by Jedlik Innovation Ltd. We thank the cytogenetic team at DPC OHII Molekuláris Genetikai Laboratórium for clinical validation, expert annotation, and reference-system access, and PPCU-ITK for research collaboration. We also acknowledge the open-source PyTorch, Detectron2, FastAPI, and TorchServe communities. Earlier versions of the clinical-validation results were presented as posters at three cytogenetics and laboratory-medicine conferences in 2025.

\section*{References}

\textbf{Cytogenetic AI:}

[You et al.~2023] You, S., Xia, J., et al.~(2023). \emph{AutoKary2022: A Large-Scale Densely Annotated Dataset for Chromosome Instance Segmentation.} IEEE ICME. arXiv:2303.15839.

[Xia et al.~2024] Xia, J., Wang, J., et al.~(2024). \emph{KaryoXpert: An accurate chromosome segmentation and classification framework.} Computers in Biology and Medicine 177:108601.

[Lin et al.~2023] Lin et al.~(2023). \emph{DaCSeg: Divide and conquer for accurate overlapping chromosome instance segmentation in metaphase cell images.} Biomedical Signal Processing and Control.

\medskip
\textbf{Foundational architectures and methods:}

[He et al.~2017] He, K., Gkioxari, G., Dollár, P., \& Girshick, R. (2017). \emph{Mask R-CNN.} ICCV 2017. arXiv:1703.06870.

[He et al.~2016] He, K., Zhang, X., Ren, S., \& Sun, J. (2016). \emph{Deep Residual Learning for Image Recognition.} CVPR 2016. arXiv:1512.03385.

[Ren et al.~2015] Ren, S., He, K., Girshick, R., \& Sun, J. (2015). \emph{Faster R-CNN: Towards Real-Time Object Detection with Region Proposal Networks.} NeurIPS 2015. arXiv:1506.01497.

[Ronneberger et al.~2015] Ronneberger, O., Fischer, P., \& Brox, T. (2015). \emph{U-Net: Convolutional Networks for Biomedical Image Segmentation.} MICCAI 2015. arXiv:1505.04597.

[Cai \& Vasconcelos 2018] Cai, Z., \& Vasconcelos, N. (2018). \emph{Cascade R-CNN: Delving into High Quality Object Detection.} CVPR 2018. arXiv:1712.00726.

[Liu et al.~2021] Liu, Z., Lin, Y., et al.~(2021). \emph{Swin Transformer: Hierarchical Vision Transformer using Shifted Windows.} ICCV 2021. arXiv:2103.14030.

[Cheng et al.~2022] Cheng, B., Misra, I., Schwing, A. G., Kirillov, A., \& Girdhar, R. (2022). \emph{Masked-attention Mask Transformer for Universal Image Segmentation.} CVPR 2022. arXiv:2112.01527.

[Otsu 1979] Otsu, N. (1979). \emph{A Threshold Selection Method from Gray-Level Histograms.} IEEE Trans.~SMC 9(1):62--66.

[Alguacil et al.~2021] Alguacil, A., et al.~(2021). \emph{Effects of boundary conditions in fully convolutional networks.} arXiv:2106.11160.

\end{document}